\pdfoutput=1

\documentclass[11pt]{article}

\usepackage[final]{acl}

\usepackage{times}
\usepackage{latexsym}
\usepackage{xcolor}

\usepackage{tcolorbox}
\usepackage[frozencache,cachedir=.]{minted}
\newtcolorbox[auto counter]{mytbox}[2][]{%
colback=blue!5!white,
colframe=blue!75!black,
fonttitle=\bfseries,
fontupper=\small,
fontlower=\small,
title=Ex.~\thetcbcounter: #2,#1}

\usepackage[T1]{fontenc}
\usepackage[utf8]{inputenc}
\usepackage[greek,english]{babel}
\usepackage{alphabeta}

\usepackage{soul}

\usepackage[utf8]{inputenc}

\usepackage{microtype}

\usepackage{inconsolata}

\usepackage{graphicx}

\title{Krikri: Advancing Open Large Language Models for Greek}

\author{Dimitris Roussis\thanks{Equal contribution}, {\bf Leon Voukoutis\footnotemark[1]}, {\bf Georgios Paraskevopoulos\footnotemark[1]}, \\ {\bf Sokratis Sofianopoulos\footnotemark[1]}, {\bf Prokopis Prokopidis\footnotemark[1]}, {\bf Vassilis Papavasileiou}, \\ {\bf Athanasios Katsamanis}, {\bf Stelios Piperidis}, {\bf Vassilis Katsouros\footnotemark[1]} \\
\\
  Institute for Speech and Language Processing, Athena Research Center\\ Artemidos 6 \& Epidavrou, Athens, Greece\\
  \texttt{vsk@athenarc.gr}}

\begin{document}
\maketitle
\begin{abstract}

We introduce Llama-Krikri-8B, a cutting-edge Large Language Model tailored for the Greek language, built on Meta’s Llama 3.1-8B. Llama-Krikri-8B has been extensively trained on high-quality Greek data to ensure superior adaptation to linguistic nuances. With 8 billion parameters, it offers advanced capabilities while maintaining efficient computational performance. Llama-Krikri-8B supports both Modern Greek and English, and is also equipped to handle polytonic text and Ancient Greek. 
The chat version of Llama-Krikri-8B features a multi-stage post-training pipeline, utilizing both human and synthetic instruction and preference data created via various synthesis techniques, such as MAGPIE, in order to address the issue of data scarcity for Greek. In addition, for evaluation, we propose three novel public benchmarks for Greek which are suitable for instruction-tuned models. Our evaluation on existing as well as the proposed benchmarks shows notable improvements over comparable Greek and multilingual LLMs in both natural language understanding and generation as well as tasks related to math, code, instruction-following, etc.
\end{abstract}

\section{Introduction}\label{sec:intro}
Recent advancements in AI have been largely driven by the development of large-scale foundation models. Meta’s Llama 3 \cite{grattafiori2024llama3herdmodels} fostered a new generation of open models, designed for strong multilingual capabilities, code generation, reasoning, and tool use. With extended context windows, and refined training strategies, models based on Llama 3 have achieved performance comparable to proprietary systems like GPT-4. A critical aspect in this evolution is the development of multilingual and language-specific models, democratizing access to AI technologies and preserving linguistic diversity.

While substantial progress has been made for widely spoken languages, low and medium resource languages remain underrepresented. Greek, in particular, has received limited attention despite its linguistic complexity, rich cultural heritage, and historical significance. Addressing this gap, we present Llama-Krikri-8B, a cutting-edge open Large Language Model tailored for the Greek language. Built on Meta’s Llama 3.1-8B architecture, Llama-Krikri has been continually pretrained on a diverse, high-quality Greek corpus. This allows the model to effectively capture the syntactic and semantic nuances of Greek, while retaining the multilingual strengths of the base model. Notably, Llama-Krikri also supports English and is capable of handling polytonic and Ancient Greek texts, addressing not only contemporary but also historical forms of the language.

Compared to Meltemi-7B \cite{voukoutis2024}, the previous state-of-the-art open Greek LLM built on Mistral 7B \cite{jiang2023mistral}, Llama-Krikri-8B significantly increases the number of parameters, context length, and training data scale. Additionally, it features an enhanced post-training pipeline using both human and synthetic data. By utilizing various data synthesis methodologies, such as MAGPIE \cite{xu2024magpiealignmentdatasynthesis}, we generate high-quality synthetic data for Supervised Fine-Tuning (SFT) and alignment via Direct Preference Optimization (DPO) \cite{rafailov2024directpreferenceoptimizationlanguage}. This pipeline addresses the scarcity of relevant data for Greek and ensures that the model produces helpful, honest, and harmless outputs.

To evaluate Llama-Krikri-8B, we also introduce three novel public benchmarks specifically designed for Greek instruction-tuned models. These, alongside existing evaluation suites, show that Llama-Krikri outperforms comparable Greek and multilingual LLMs in both natural language understanding and generation, as well as other challenging tasks related to math, code, instruction-following, etc. Moreover, it supports function calling and agentic behavior, opening new application domains for Greek users. Llama-Krikri-8B is available under the Llama 3.1 Community License Agreement\footnote{\url{https://www.llama.com/llama3_1/license/}}.

Our key contributions are:
\begin{itemize}
    \item We present Llama-Krikri-8B, a state-of-the-art open Greek foundation model based on Llama 3.1, demonstrating strong capabilities in Modern and Ancient Greek, math, code, instruction-following, etc., while also supporting function calling and agentic behavior.
    \item We incorporate additional steps aiming to increase efficiency, reduce training costs, and improve performance, such as expanding the original Llama 3.1 tokenizer and embeddings, and conducting an annealing phase with high-quality and synthetic data. 
    \item We collect and curate a large pretraining corpus and utilize synthetic data creation strategies for post-training, in order to address the scarcity of Greek data.
    \item We implement a multi-step post-training pipeline incorporating Supervised Fine-Tuning (SFT) and alignment through Direct Preference Optimization (DPO), which leads to Llama-Krikri-8B-Instruct surpassing Llama-3.1-8B-Instruct in several English tasks.
    \item We introduce three new benchmarks for evaluating instruction-tuned Greek LLMs, covering instruction-following, multi-turn chats, code/math tasks, etc.
    \item We show that Llama-Krikri-8B and Llama-Krikri-8B-Instruct significantly outperform existing open multilingual models in Greek tasks across several domains, while also being competitive with 3-4 times larger models.
\end{itemize}

\section{Background and Related Work}\label{sec:background}

Large Language Models (LLMs) have achieved state-of-the-art performance across a wide variety of natural language processing (NLP) tasks. These models are typically trained on massive corpora dominated by English, leading to strong performance in English-language tasks but comparatively weaker capabilities in other languages \cite{devlin2019bertpretrainingdeepbidirectional, brown2020languagemodelsfewshotlearners}. As a result, the development of language-specific LLMs has become an active area of research, particularly for under-represented languages.

One prominent strategy for developing such models is continual pretraining, where a pretrained base model is further trained on data in the target language. This approach allows researchers to leverage the general capabilities of large base models while improving performance in specific linguistic domains, without the prohibitive cost of training from scratch \cite{gururangan2020dontstoppretrainingadapt}.

Several recent projects have successfully applied continual pretraining to adapt existing models to new languages. BgGPT-GEMMA-2-27B-Instruct \cite{alexandrov2024bggpt10extendingenglishcentric} fine-tunes Google’s Gemma-2 model~\cite{gemmateam2024gemma2improvingopen} for Bulgarian, combining over 100B tokens of Bulgarian and English data and applying techniques such as Branch-and-Merge to mitigate catastrophic forgetting. Similarly, LeoLM \cite{Pluster2023} adapts a Llama-based architecture for German, while the Sabiá \cite{10.1007/978-3-031-45392-2_15} and Bielik models \cite{ociepa2024bielik7bv01polish} adapt Mistral-based architectures for Portuguese and Polish respectively, through targeted continual pretraining and post-training.

For Greek, Meltemi-7B represents the first open generative LLM tailored to the language \cite{voukoutis2024}. It was developed by continually pretraining Mistral-7B on a substantial Greek corpus, followed by instruction fine-tuning. While effective, Meltemi's performance is bounded by the size and capabilities of the base model, data scarcity for instruction tuning in Greek, as well as the limited post-training alignment techniques employed at the time.

Beyond language adaptation, alignment of LLMs to generate helpful, harmless, and honest outputs has become increasingly central. Early approaches such as InstructGPT \cite{ouyang2022traininglanguagemodelsfollow} and Constitutional AI \cite{bai2022constitutionalaiharmlessnessai} rely on multi-stage fine-tuning pipelines involving human feedback or rule-based constraints. More recently, DPO and data synthesis methods like MAGPIE have enabled scalable and effective post-training. MAGPIE, in particular, leverages already-aligned models to generate large volumes of instruction-response pairs, demonstrating that synthetic data can rival or surpass human-curated datasets.

These advancements highlight a trend toward bootstrapping high-quality training data using strong base models, especially in low-resource languages. Our work builds on this foundation by employing Llama 3.1 as a base architecture, employing various data synthesis techniques (e.g., MAGPIE), and applying a more rigorous post-training pipeline, including DPO for alignment.

\begin{table*}[t]
  \centering
\begin{tabular}{ccccc}
\hline
 & \textbf{Original} & & \textbf{Upsampled} & \\
\textbf{Subcorpus} & \textbf{Tokens (B)} & \textbf{Percentage} & \textbf{Tokens (B)} & \textbf{Percentage} \\
\hline
Greek & 56.7 & 62.3\% & 66.1 & 60.0\% \\
English & 21.0 & 23.1\% & 25.2 & 22.9\% \\
Parallel & 5.5 & 6.0\% & 8.8 & 8.0\% \\
Math/Code & 7.8 & 8.6\% & 10.1 & 9.1\% \\
\hline
\textbf{Total} & \textbf{91.0} & \textbf{100\%} & \textbf{110.2} & \textbf{100\%} \\
\hline
\end{tabular}

\caption{Composition of the pretraining corpus - original and upsampled}
\label{table:pretraining-data}

\end{table*}

\section{Methodology}\label{sec:method} 

Llama-Krikri-8B is based on the Transformer architecture \cite{vaswani2023attentionneed}, which has become the de facto standard for large language models. The model inherits its architecture from Meta’s Llama 3.1-8B, leveraging the strong foundation in multilingual understanding, code generation, and reasoning provided by Llama 3.1. 

Adapting an LLM for the Greek language requires addressing the lack of high-quality Greek data in the massive datasets typically used to train foundation models. Even though Llama 3.1’s pretraining corpus comprises 
trillions of tokens, it struggles to generate coherent Greek text, thus indicating that Greek data is only a tiny fraction of its training data; we should note however that there is limited information on the composition of its pretraining data \cite{grattafiori2024llama3herdmodels}. Our approach is to perform continual pretraining with Greek and parallel data to infuse the model with Greek knowledge. This training must be done carefully to avoid catastrophic forgetting \cite{luo2025empiricalstudycatastrophicforgetting} of the base model’s prior knowledge in other languages and domains, and it must also be followed by post-training in order to ensure that the model follows instructions adequately and safely.

In the subsections below, we detail our comprehensive methodology for developing Llama-Krikri-8B. We begin by describing our process for collecting, filtering, and cleaning a diverse corpus of Greek text supplemented with English, parallel, and domain-specific content (\S\ref{sec:corpus}). Next, we explain how we expanded the tokenizer to better handle Greek text (\S\ref{sec:tokenizer}) and initialized the corresponding embeddings (\S\ref{sec:embedding_pretrain_process}). We then detail our continual pretraining procedure that extends the base model while preserving its original capabilities (\S\ref{sec:pretrain_process}), followed by an annealing phase using high-quality curated data to further refine performance (\S\ref{sec:annealing}). Finally, we describe our instruction tuning pipeline, which includes supervised fine-tuning and preference optimization to align the model with human values (\S\ref{sec:instruction_tuning}). This comprehensive approach addresses both the technical challenges of adapting a multilingual foundation model for Greek while ensuring the resulting model maintains its general capabilities and even outperforms the instruction-tuned variant of Llama-3.1-8B across several tasks.

\subsection{Pretraining Data Collection \& Cleaning}\label{sec:corpus}
As a foundation for continual pretraining, we curated a large corpus of texts totalling approximately 91 billion tokens (after filtering and deduplication), which was upsampled to 110 billion tokens for the final pretraining mix. This corpus was constructed with a primary focus on Greek by utilizing sources such as Wikipedia\footnote{\scriptsize{\url{https://huggingface.co/datasets/wikimedia/wikipedia}}}, Wikisource\footnote{\scriptsize{\url{https://huggingface.co/datasets/wikimedia/wikisource}}}, Project Gutenberg\footnote{\scriptsize{\url{https://www.gutenberg.org/}}}, governmental documents, academic repositories, literature, and other openly available datasets. We also included Ancient Greek texts (as well as other variants of Greek), high-quality English texts, parallel datasets in language pairs encompassing Greek, English, French, German, Portuguese, Italian, and Spanish, as well as datasets containing math and code, so as to retain or even improve the model's existing capabilities.
The distribution included 56.7 billion monolingual Greek tokens (62.3\%), 21 billion monolingual English tokens (23.1\%), 5.5 billion parallel data tokens (6.0\%), and 7.8 billion math and code tokens (8.6\%).
Table \ref{table:pretraining-data} presents the distribution of the pretraining data mix, with more details provided in Appendix \ref{sec:appendix:data_mix} \nameref{sec:appendix:data_mix}.

After corpus collection, we implemented a multi-stage preprocessing and filtering pipeline to ensure a high quality for the pretraining data. Various parts of our filtering methodology have been informed by approaches used in previous work 
\citet{voukoutis2024} and large-scale corpus creation efforts such as Zyda \cite{tokpanov2024zyda}. However, we have adapted these approaches to cater for the peculiarities of the Greek language, as well as different data types, including PDF extraction for academic and legal texts, rule-based filtering for web content, and model-based quality assessment for parallel data using LASER \cite{artetxe2018margin,artetxe2019massively} and CometKiwi \cite{rei-etal-2022-cometkiwi}. Other cleaning approaches aim for privacy protection via the anonymization of personally identifiable information, for increased training efficiency via deduplication, and for assigning higher importance to important or more educational content (e.g., legal, scientific, medical, translation) via upsampling \cite{carlini2023quantifying,txt360data2024}. We detail the preprocessing pipelines we used in Appendix \ref{sec:appendix:data_cleaning} \nameref{sec:appendix:data_cleaning}.

\subsection{Tokenizer Expansion}\label{sec:tokenizer}
The original Llama 3 tokenizer comprises $128,000$ tokens and is  inefficient for Greek texts, as it generally performs character-level tokenization for Greek. This was determined, through the approximation of the Llama-3.1 tokenizer's fertility \cite{csaki2023efficiently}, a metric of the average tokens per word produced. To determine the efficiency of the original Llama 3 tokenizer and compare with our approach, we conducted tests on diverse Greek and English corpora (each one containing $100,000$ rows and totalling approximately 2M words) and calculated the difference in fertility, as can be seen in Table \ref{table:tokenizers}. We observe that the Llama-3.1-8B tokenizer exhibits a fertility of 2.73 for Greek and 1.33 for English. Our Llama-Krikri-8B tokenizer demonstrates a significantly lower fertility of 1.65 for Greek, while maintaining the same low fertility of 1.33 for English as the base Llama-3.1-8B tokenizer.

In order to develop an optimal tokenizer for Greek which is also efficient in historical dialects of the language, as well as in critical domains, we extended the Llama 3 tokenizer with 20,992 new tokens through a multi-stage process which encompasses curating high-quality texts and allocating new tokens across five domains: (1) General, (2) Legal, (3) Scientific, (4) Literature, and (5) Ancient Greek. This process is especially important during model inference, as it significantly reduces the input and output token cost during model use. Furthermore, more compact representations of input text help to improve model performance.
We provide details on the steps for the tokenizer in Appendix \ref{sec:appendix:tokenizer-expansion-details}, \nameref{sec:appendix:tokenizer-expansion-details}, which also details the data construction process for creating the tokenizer with different corpora than the ones we used to evaluate the tokenizer fertilities presented above.

\begin{table}[ht]
  \centering
  \small
\begin{tabular}{lccc}
\hline
& \textbf{Vocabulary} & \textbf{Fertility} & \textbf{Fertility} \\
\textbf{Tokenizer} & \textbf{Size} & \textbf{Greek} & \textbf{English}\\
\hline
Mistral-7B & 32,000 & 6.80 & 1.49 \\
Meltemi-7B & 61,362 & \textbf{1.52} & 1.44 \\
Llama-3.1-8B & 128,000 & 2.73 & \textbf{1.33} \\
Llama-Krikri-8B & 149,248 & 1.65 & \textbf{1.33} \\
\hline
\end{tabular}
\caption{Tokenizer statistics for Greek and English}
\label{table:tokenizers}
\vspace{-1em}
\end{table}

\subsection{Greek Embeddings Expansion \& Training}\label{sec:embedding_pretrain_process}
To effectively integrate newly introduced Greek tokens into the model, we implemented an initial, targeted training phase for their corresponding embeddings. By preparing the new token representations prior to full-scale pretraining, we mitigate potential disruptions to the existing model parameters. 

We initialized the model with Llama 3.1‑8B-Base weights, freezing all but the embeddings and output‐projection weights for the 20,992 new tokens, allowing their initial training without large gradient updates to the rest of the model. Rather than random initialization, we calculated each new token's initial embedding by averaging the embeddings associated with the tokens that would have been produced by the original Llama-3.1 tokenizer when processing that token. The dataset for this step was comprised of 5B tokens and was sampled to be representative of the overall pre-training corpus. This short, several-thousand-step training regimen ensured a smoother integration of the new vocabulary into the model's existing knowledge representation. More details are described in Appendix \ref{sec:appendix:embedding_training}, \nameref{sec:appendix:embedding_training}.

\subsection{Continual Pretraining Process}\label{sec:pretrain_process}
After embedding training, all parameters were unfrozen, and training continued on the 110B token corpus using a mixed-curriculum strategy, employing a dataset sampling schedule that prefers data closer to the initial Llama-3.1 distribution in the beginning, while shifting closer to our true dataset distribution as training continues, and re-warming and re-decaying the learning rate \cite{ibrahim2024simple}. 

We utilized a curriculum learning strategy which evolved throughout the training process, beginning with a higher proportion of simpler, high-quality Greek texts and English content to establish foundational Greek language skills while maintaining the model's original capabilities. As training progressed, we gradually introduced more diverse and complex Greek sources while decreasing the relative proportion of English content. Additionally, we periodically introduced small "replay" segments of high-quality English or code to refresh the model's memory of content it might not frequently encounter in the Greek-focused corpus. This approach is similar to techniques observed by \citet{alexandrov2024bggpt10extendingenglishcentric}, who emphasized careful data ordering to preserve base model skills when adapting to another language.

Training was conducted over $\sim$50 days on the 110B token dataset at 128K context length, on 2 nodes of 8 NVIDIA H200 GPUs, using Zero 3 for parallelization and bf16 mixed precision for computational efficiency. The detailed configuration of the pretraining procedure can be found in Appendix \ref{sec:appendix:pretrain_process}, \nameref{sec:appendix:pretrain_process}.

\subsection{Annealing Phase}\label{sec:annealing}
Following pretraining, a short annealing pass used a curated 3.5B token dataset of very high-quality texts across all subcorpora. We used within-dataset normalized perplexity, calculated using 5-gram KenLM~\cite{heafield2011kenlm}, to implement a dataset-aware fluency scoring method for document selection. This approach aimed at preserving dataset diversity while removing intra-dataset outliers with lower linguistic coherence. To boost comprehension and reasoning, a synthetic question-answer dataset (189M tokens) was created by curating a high-quality corpus and prompting stronger models to generate Q\&A triplets with reasoning from curated documents. Ablations revealed that adding this synthetic QA component significantly enhanced performance, not only improving Greek metrics (+2.1 points over continual pretraining alone) but notably restoring and even improving English capabilities beyond the original Llama-3.1-8B (+0.8 points). These ablations as well as the training configuration for the annealing phase can be found in Appendix \ref{sec:appendix:annealing}, \nameref{sec:appendix:annealing}.

\subsection{Instruction Tuning and Alignment}\label{sec:instruction_tuning}
Llama-Krikri-8B-Instruct was created by fine-tuning the base model for instruction following and dialogue. In order to address Greek data scarcity, the pipeline combined data synthesis, filtering, two-stage Supervised Fine-Tuning (SFT), and Direct Preference Optimization (DPO).

Data collection, synthesis, \& curation involved collecting datasets with high-quality:
\begin{itemize}
    \item Instruction pairs, sourced from Tulu 3\footnote{\scriptsize{\url{https://huggingface.co/collections/allenai/tulu-3-datasets-673b8df14442393f7213f372}}} \cite{lambert2025tulu3pushingfrontiers}, SmolTalk\footnote{\scriptsize{\url{https://huggingface.co/datasets/HuggingFaceTB/smoltalk}}} \cite{allal2025smollm2smolgoesbig}, AgentInstruct\footnote{\scriptsize{\url{https://huggingface.co/datasets/microsoft/orca-agentinstruct-1M-v1}}} \cite{mitra2024agentinstructgenerativeteachingagentic}, MAGPIE Ultra v1.0\footnote{\scriptsize{\url{https://huggingface.co/datasets/argilla/magpie-ultra-v1.0}}} \cite{xu2024magpiealignmentdatasynthesis}, etc.
    \item Preference triplets, sourced from UltraFeedback\footnote{\scriptsize{\url{https://huggingface.co/datasets/openbmb/UltraFeedback}}} \cite{cui2023ultrafeedback}, UltraMedical\footnote{\scriptsize{\url{https://huggingface.co/datasets/TsinghuaC3I/UltraMedical-Preference}}} \cite{zhang2024ultramedical}, UltraInteract\footnote{\scriptsize{\url{https://huggingface.co/datasets/openbmb/UltraInteract_pair}}} \cite{yuan2024advancing}, HelpSteer2\footnote{\scriptsize{\url{https://huggingface.co/datasets/nvidia/HelpSteer2}}} \cite{wang2024helpsteer2,wang2024helpsteer2preferencecomplementingratingspreferences},  WildGuard\footnote{\scriptsize{\url{https://huggingface.co/allenai/wildguard}}} \cite{wildguard2024}, etc.
\end{itemize}

Additionally, Greek data was synthesized via translation (with post-editing) of several of the aforementioned datasets, regenerating responses using LLMs such as Gemma-2-27B-IT \cite{gemmateam2024gemma2improvingopen}, and generating synthetic instructions directly in Greek using the MAGPIE technique (steering the model with system templates towards general, reasoning, and natural language processing tasks). Curated corpora from the annealing phase were reused for synthetic Q\&As and multi-turn dialogues created through few-shot prompting techniques. Throughout this process, we utilized the Skyword-Reward-Gemma-2-27B-v0.2~\cite{liu2024skywork} reward model in order to score and filter the data, as it demonstrates high accuracy in benchmarks such as RewardBench \cite{lambert2024rewardbenchevaluatingrewardmodels}. Several processing steps and rule-based filters were also applied across all datasets so as to validate dialogue structure, ensure consistent formatting, verify language, and remove empty system messages, special tokens, and other artifacts.

SFT was done in two stages ($\sim$856k pairs in Stage 1, $\sim$638k in Stage 2), with progressively higher data quality. Datasets included filtered original English data, reward-model-filtered synthetic MAGPIE data (higher scores in Stage 2), translated/post-edited data (Stage 1), regenerated responses (including a "thinking" section in Stage 2), multi-language translation data, synthetic QA, synthetic multi-turn dialogues, and upsampled manual safety data. SFT produced a strong model that followed instructions but needed alignment for helpfulness, precision, and safety.

DPO provided final alignment using $\sim$92k preference triplets. Data included a selection of high-scored original and translated preferences, preferences from MAGPIE-synthesized data and regenerated translated data using reward model scores of multiple candidate responses, preferences derived via contrasting regenerated vs. translated references which aim to mitigate issues introduced by unwanted translation artifacts \cite{dang2024rlhf, dang2024ayaexpansecombiningresearch}, and safety preferences including manually created data targeting Greek-specific safety concerns and sensitivities. 

DPO maximized the likelihood of preferred responses while minimizing that of dispreferred ones, while an added length normalization term mitigated the empirical phenomenon of DPO disproportionally preferring longer sequences. DPO significantly improved response quality, safety, and helpfulness compared to the SFT-only model, and resulted to our final model, i.e., Llama-Krikri-8B-Instruct. Implementation details for both SFT and DPO can be found in Appendix \ref{sec:appendix:instruction_tuning}, \nameref{sec:appendix:instruction_tuning}.

\begin{table*}[h!]
\centering
\begin{tabular}{lccc}
\hline
\textbf{Benchmark} & \textbf{Meltemi-7B-v1.5} & \textbf{Llama-3.1-8B} & \textbf{Krikri-8B-Base} \\
\hline
Medical MCQA EL (15-shot) & 42.2 & 33.4 & \textbf{53.8} \\
Belebele EL (5-shot) & 61.0 & 72.8 & \textbf{82.7} \\
HellaSwag EL (10-shot) & 53.8 & 52.1 & \textbf{64.6} \\
ARC-Challenge EL (25-shot) & 40.0 & 39.9 & \textbf{49.4} \\
TruthfulQA MC2 EL (0-shot) & 49.0 & 51.1 & \textbf{54.2} \\
MMLU EL (5-shot) & 41.2 & 42.6 & \textbf{52.0} \\
\hline
Average & 47.9 & 48.7 & \textbf{59.5} \\
\hline
\end{tabular}
\caption{Greek benchmark results (accuracy \%) for base models.}
\label{table:greek_results}
\end{table*}

\section{Evaluation}\label{sec:evaluation}

In this section, we present evaluation details for Llama-Krikri-8B-Base and Llama-Krikri-8B-Instruct, across six Greek and six English benchmarks. 
We compare our base model directly with the base model 
Llama-3.1-8B \cite{grattafiori2024llama3herdmodels} and the previous Greek state-of-the-art model Meltemi-7B-v1.5 \cite{voukoutis2024}. Additionally, we evaluate our chat model, Llama-Krikri-8B-Instruct on three challenging English benchmarks, as well as three novelly constructed Greek benchmarks which correspond to the English ones.

\subsection{Base Model Evaluation: Krikri-8B-Base}\label{sec:evaluation_base}

We evaluated Llama-Krikri-8B-Base against Llama-3.1-8B  and Meltemi-7B-v1.5  in a few-shot setting, consistent with the Open LLM Leaderboard\footnote{\scriptsize{\url{https://huggingface.co/spaces/open-llm-leaderboard/open_llm_leaderboard}}}.

\paragraph{Greek Benchmarks:}
The evaluation was carried out on a suite of six Greek-specific benchmarks\footnote{\scriptsize{\url{https://huggingface.co/collections/ilsp/ilsp-greek-evaluation-suite-6827304d5bf8b70d0346b02c}}} used in \citet{voukoutis2024}, including machine-translated versions of established English datasets (ARC-Challenge Greek, Truthful QA Greek, HellaSwag Greek, MMLU Greek
), the existing Belebele Greek benchmark~\cite{bandarkar2024belebele}, and a novel medical QA benchmark (Medical MCQA).


Results in Table \ref{table:greek_results} demonstrate substantial improvements for Greek (+10.8\%) compared to  Llama-3.1-8B.  Moreover, we observe that Llama-Krikri-8B-Base surpasses  Meltemi-7B-v1.5 with a notable +11.6\% average improvement across all benchmarks. On MMLU Greek, Llama-Krikri-8B-Base surpasses Llama-3.1-8B and Meltemi-7B-v1.5 by +9.4\% and +10.8\% respectively, while on ARC-Challenge Greek, it achieves an accuracy of 49.4\%, compared to Llama-3.1-8B's and Meltemi-7B-v1.5's 39.9\% and 40.0\%, respectively. Similar substantial gains are observed on the Belebele Greek dataset, where Llama-Krikri-8B-Base scores 82.7\%, surpassing Meltemi-7B-v1.5 and Llama-3.1-8B by +21.7\% and +9.9\%, respectively. In the Greek Medical MCQA, Llama-Krikri-8B-Base reaches 53.8\%, demonstrating clear advancements over Llama-3.1-8B  (+20.4\%) in a domain-specific Greek benchmark that was not translated from English.


\begin{table*}[h!]
\centering
\begin{tabular}{lccc}
\hline
\textbf{Benchmark} & \textbf{Meltemi-7B-v1.5} & \textbf{Llama-3.1-8B} & \textbf{Krikri-8B-Base} \\
\hline
Winogrande (5-shot) & 73.4 & \textbf{74.6} & 72.6 \\
Belebele EN (5-shot) & 77.7 & 71.5 & \textbf{79.8} \\
HellaSwag EN (10-shot) & 79.6 & \textbf{82.0} & 80.7 \\
ARC-Challenge EN (25-shot) & 54.1 & \textbf{58.5} & 57.8 \\
TruthfulQA MC2 EN (0-shot) & 40.5 & 44.2 & \textbf{44.8} \\
MMLU EN (5-shot) & 56.9 & \textbf{66.2} & 65.1 \\
\hline
Average & 63.7 & 66.2 & \textbf{67.0} \\
\hline
\end{tabular}
\caption{English benchmark results (accuracy \%) for base models.}
\label{table:english_results_base}
\end{table*}

\paragraph{English Benchmarks}
For the evaluation of base models on English, we utilized six benchmarks, with five of them being the original versions of those also used for Greek: ARC-Challenge~\cite{allenai:arc}, Truthful QA~\cite{lin-etal-2022-truthfulqa}, HellaSwag~\cite{zellers-etal-2019-hellaswag}, MMLU~\cite{hendrycks2021measuring}, and Belebele~\cite{bandarkar2024belebele}. Additionally, the Winogrande~\cite{sakaguchi2021winogrande} test set was used as the sixth benchmark for English.  In the results presented in Table~\ref{table:english_results_base} we see that our training methodology not only mitigates catastrophic forgetting effectively, but also improves average performance across all English test sets by +0.8\%.  

\subsection{Chat Model Evaluation: Krikri-8B-Instruct}

For evaluating the capabilities of Llama-Krikri-8B-Instruct as a conversational assistant, suitable for multi-turn dialogue, instruction-following and complex coding and math queries, we used a suite of benchmarks in both English and Greek. For English, we conducted evaluations across two paths:
\begin{itemize}
    \item We submitted our model to the Open LLM Leaderboard \cite{open-llm-leaderboard-v2} which automatically evaluates models on IFEval, BBH, MATH, GPQA, MUSR, and MMLU-Pro using the Eleuther AI Language Model Evaluation Harness \cite{eval-harness}, a unified framework to test generative language models on a large number of different evaluation tasks.
    \item We used the Arena Hard Auto v0.1 \cite{li2024crowdsourced,chiang2024chatbot}, IFEval \cite{zhou2023instruction} (strict avg) and MT-Bench \cite{zheng2023judging} benchmarks. Although IFEval was already included in the Open LLM Leaderboard, we re-implemented it to enable accurate comparison with multiple models. In the evaluation of MT-Bench we used GPT-4o (2024-08-06) as the judge model, while in the evaluation of Arena Hard Auto v0.1 we used the standard approach with GPT-4-0314 as the baseline model (by default scoring 50\%) and GPT-4-1106-Preview as the judge model, while also reusing the generations and judgments already computed by the authors.
\end{itemize}

\begin{table*}[t!]
\centering
\begin{tabular}{lcccc}
\hline
\textbf{Model} & \textbf{IFEval EL} & \textbf{IFEval EN} & \textbf{MT-Bench EL} & \textbf{MT-Bench EN} \\
\hline
Qwen 2.5 7B & 46.2 & 74.8 & 5.83 & \textbf{7.87} \\
EuroLLM 9B & 51.3 & 64.5 & 5.98 & 6.27 \\
Aya Expanse 8B & 50.4 & 62.2 & 7.68 & 6.92 \\
Meltemi-7B-v1.5 & 32.7 & 41.2 & 6.25 & 5.46 \\
Llama-3.1-8B & 45.8 & 75.1 & 6.46 & 7.25 \\
Llama-Krikri-8B & \textbf{67.5} & \textbf{82.4} & \textbf{7.96} & 7.21 \\
\hline
Gemma 2 27B IT & 63.2 & 75.6 & 8.23 & \textbf{8.00} \\
Aya Expanse 32B & 60.3 & 70.2 & \textbf{8.27} & 7.40 \\
\hline
\end{tabular}
\caption{Greek and English evaluation results using IFEval and MT-Bench.}
\label{table:chat_eval_results}
\end{table*}

For Greek, we created three novel evaluation benchmarks by translating three challenging, diverse, and widely used English benchmarks, ensuring high-quality translations through careful post-editing and validation:
\begin{itemize}
    \item \textbf{IFEval Greek}\footnote{\scriptsize{\url{https://huggingface.co/datasets/ilsp/ifeval_greek}}} 
    (strict avg.): a manual translation of 541 prompts from the original Instruction-Following Evaluation benchmark \cite{zhou2023instruction}, featuring verifiable instructions such as "απάντησε με περισσότερες από 400 λέξεις" (answer with more than 400 words) and "ανάφερε τη λέξη ΤΝ τουλάχιστον 3 φορές" (mention the word AI at least 3 times), designed to assess the model's ability to follow specific instructions. 
    \item \textbf{MT-Bench Greek}\footnote{\scriptsize{\url{https://huggingface.co/datasets/ilsp/mt-bench-greek}}},
    a translated version of the Multi-turn Benchmark \cite{zheng2023judging} containing 80 high-quality, multi-turn conversations across eight diverse categories (e.g., STEM, humanities, roleplay, coding, etc.), carefully post-edited to ensure natural Greek phrasing and cultural appropriateness. MT-Bench is also used to evaluate the function-calling capabilities of LLMs~\cite{chen-etal-2025-enhancing}. The performance of each model is calculated using LLM-as-Judge \cite{zheng2023judging} with GPT-4o (2024-08-06) serving as the scoring model. 
    \item \textbf{Arena-Hard-Auto Greek}\footnote{\scriptsize{\url{https://huggingface.co/datasets/ilsp/m-ArenaHard_greek}}}, a translated version of Arena-Hard-Auto v0.1, which originates from Chatbot Arena \cite{chiang2024chatbot} was included in m-ArenaHard~\cite{dang2024ayaexpansecombiningresearch} after translation with Google Translate API v3. We later post-edited using Claude Sonnet 3.5 \cite{anthropic2024claude} with 10-shot examples to address translation issues, particularly in coding-related prompts where some parts would best be left untranslated, as well as to retain the original style of the prompts, since some of them would be best left vaguely posed as in the original prompt. We used the version of the benchmark with style control methods for Markdown elements\footnote{\scriptsize{\url{https://lmsys.org/blog/2024-08-28-style-control/}}}. We set GPT-4o-Mini (2024-07-18) as the baseline model (by default 50\% score) and GPT-4o (2024-08-06) as the judge model.
\end{itemize}

As shown in Table \ref{table:chat_eval_results}, Llama-Krikri-8B-Instruct demonstrates exceptional performance across both Greek and English benchmarks, substantially outperforming not only its parent model Llama-3.1-8B-Instruct but also other competitive multilingual models in the 7-9B parameter range. It should be noted that the IFEval scores reported in this table reflect our own implementation of the benchmark, which may differ from the Open LLM Leaderboard implementation due to variations in prompt formatting and evaluation criteria. Despite these methodological differences, the relative performance comparisons remain valid within each implementation context.

On IFEval Greek, Llama-Krikri-8B-Instruct achieves a remarkable 67.5\% accuracy, surpassing Llama-3.1-8B-Instruct by +21.7\% and Meltemi-7B-v1.5 by +34.8\%. Notably, our 8B model even outperforms much larger models like Gemma 2 27B IT (+4.3\%) and Aya Expanse 32B (+7.2\%) on this Greek instruction-following benchmark. As regards the original English IFEval, Llama-Krikri-8B-Instruct scores 82.4\%, significantly higher than all other models, including those with 3-4 times more parameters. This dramatic improvement suggests that our data synthesis and instruction tuning approach successfully addresses the unique challenges of following instructions in Greek, where naive translations of instruction data often fail to capture language-specific nuances.

For MT-Bench Greek, which evaluates multi-turn conversation quality, Llama-Krikri-8B-Instruct achieves a score of \textbf{7.96}, making it the top performer amongst other models in its size class. While larger models like Gemma 2 27B IT (8.23) and Aya Expanse 32B (8.27) achieve slightly higher scores on MT-Bench Greek, the margin is surprisingly small given the substantial difference in model size. On MT-Bench English, Llama-Krikri-8B maintains competitive performance at 7.21, essentially identical with Llama-3.1-8B-Instruct (-0.04), though understandably lower than the larger Gemma 2 27B IT (-0.79) and Aya Expanse (-0.19).

\begin{table}[t]
\centering
\small
\begin{tabular}{lcc}
\hline
& \textbf{Llama-3.1-8B} & \textbf{Krikri-8B} \\
\textbf{Benchmark} & \textbf{Instruct} & \textbf{Instruct} \\
\hline
IFEval & 49.22 & \textbf{60.79} \\
BBH & \textbf{29.38} & 29.31 \\
MATH & \textbf{15.56} & 11.78 \\
GPQA & \textbf{8.72} & 7.05 \\
MUSR & 8.61 & \textbf{10.46} \\
MMLU-PRO & \textbf{31.09} & 25.70 \\
\hline
Avg. & 23.76 & \textbf{24.18} \\
\hline
\end{tabular}
\caption{Comparative evaluation on English benchmarks from the Open LLM Leaderboard.}
\label{table:english_results_leaderboard}
\end{table}

As detailed in Table \ref{table:english_results_leaderboard}, Llama-Krikri-8B-Instruct's official Open LLM Leaderboard submission shows an average score of 24.18\% across all tests, slightly surpassing the 23.76\% of Llama-3.1-8B-Instruct. The model shows particularly impressive gains on IFEval implementation (60.79\% vs. 49.22\%) and MUSR (10.46\% vs. 8.61\%), while closely matching performance on the Big Bench Hard (BBH) benchmark (29.31\% vs. 29.38\%). Although Llama-Krikri performs slightly below Meta-Llama-3.1-8B-Instruct in the MMLU-PRO category (25.70\% vs. 31.09\%), the overall performance indicates successful retention of English capabilities during the Greek-focused continual pretraining.

\begin{table}[t]
\centering
\small
\begin{tabular}{lcc}
\hline
& \textbf{ArenaHard } & \textbf{ArenaHard} \\
\textbf{Model} & \textbf{Greek} & \textbf{English} \\
\hline
Aya Expanse 8B & 23.8 & -- \\
Llama 3.1 8B Instr. & 4.0 & 19.7 \\
Krikri 8B Instr. & \textbf{31.8} & \textbf{35.1} \\
\hline
Aya Expanse 32B & 40.1 & 45.1 \\
Gemma 2 27B IT & 32.2 & 49.6 \\
Llama 3.1 70B Instr. & 27.4 & 53.9 \\
\hline
GPT 4o Mini & 50.0 & 65.0 \\
\hline
\end{tabular}
\caption{Arena Hard evaluation results (\% win rate) for Greek and English.}
\label{table:arena_hard_results}
\vspace{-2em}
\end{table}

The results from our Arena Hard evaluations, presented in Table \ref{table:arena_hard_results}, reveal that, in the 8B parameter range, Llama-Krikri-8B-Instruct significantly outperforms its competitors, achieving a 31.8\% win rate on Arena Hard Greek compared to Aya Expanse 8B's 23.8\% and Llama 3.1 8B Instruct's 4.0\% (+27.8\% improvement). This demonstrates the effectiveness of our Greek-focused training approach. Even more impressively, Llama-Krikri-8B-Instruct achieves a 35.1\% win rate on Arena Hard English, substantially outperforming the original Llama-3.1-8B-Instruct (19.7\%) by +16.2\%, despite our focus on Greek capabilities. While Aya Expanse 32B leads on Arena Hard Greek with 40.1\%, our 8B model is on par with Gemma 2 27B IT (31.8\% vs. 32.2\%) and outperforms the 8.75 times larger Llama-3.1-70B-Instruct (27.4\%) by +4.4\% on the Greek evaluation data.

Please note that while all models trail behind GPT-4o-Mini (used as baseline on the Greek Arena Hard), recent research \cite{li2025preferenceleakagecontaminationproblem} has shown that judge models are biased towards student models, i.e., models finetuned on distilled data from the stronger \& larger teacher model which also acts as a judge. While details on the post-training data of GPT-4o-Mini are undisclosed, it would be very reasonable to assume that it has been trained -at least partly- with GPT-4 and GPT-4o serving as teacher models and, therefore, that the judges that we utilized are biased towards it compared to all other evaluated models.

This performance comparison with much larger models highlights the efficiency of our approach since Llama-Krikri-8B-Instruct achieves comparable or even superior performance on Greek benchmarks compared to models with 3-4x more parameters, while maintaining strong English capabilities. This efficiency is particularly important for deployment scenarios where computational resources may be limited, demonstrating that a carefully trained smaller model can rival much larger ones for specific languages, such as Greek, and that data synthesis and curation techniques can lead to a competitive model for English benchmarks with a much more compact approach than the one used in Llama-3.1-8B-Instruct \cite{grattafiori2024llama3herdmodels}.

Apart from the comparative evaluations mentioned above, we have performed zero-shot machine translation experiments on an Ancient-Modern Greek (grc$\leftrightarrow$ell) translation dataset\footnote{\scriptsize{\url{https://huggingface.co/datasets/ilsp/ancient-modern_greek_translations}}} that includes 100 sentences of Ancient Greek texts manually translated into Modern Greek. Using Llama-Krikri-8B-Instruct we have observed a 54.66 BLEU score for the Ancient to Modern Greek (grc$\rightarrow$ell) translation direction, with the reverse direction (ell$\rightarrow$grc) being more challenging (20.41 BLEU). 

\section{Discussion and Conclusions}\label{sec:conclusions}

In this paper, we presented Llama-Krikri-8B, a new LLM 
that exhibits significant skills in understanding and generating Greek, while also showing highly accurate handling of text in English and historical Greek dialects. We achieved this by developing an efficient tokenizer that exhibits a low token/words fertility for Greek and by further training Llama 3.1-8B using a carefully constructed dataset that covered a wide variety of domains. In evaluation experiments on a benchmark suite comprising Greek and English datasets,
we have observed that Llama-Krikri-8B performs significantly better in Greek (+10.8\%) compared to its base model, while also showing gains in English (+0.8\%). We then created Llama-Krikri-8B-Instruct, a version designed for following instructions and engaging in helpful conversations. This involved a multi-step process that comprised synthetic data generation in a multitude of domains, fine-tuning the model and then aligning it with human preferences. Evaluations revealed that Llama-Krikri-8B-Instruct significantly outperformed Llama-3.1-8B-Instruct in both Greek (+21.7\%) and English (+7.3\%) IFEval. Our model also demonstrated highly competitive chat abilities in both languages across several benchmarks.

\section{Limitations}\label{sec:limitations}

The quality and accessibility of Greek datasets are critical to the development of Krikri. Greek open-source corpora are becoming more numerous, but they might not be as large or varied as datasets for more extensively spoken languages, like English. This may result in biases in the model's understanding of Greek, especially with regard to regional variances, dialects, and specialized fields like technical fields, law, or medicine.

As an 8B parameter model, our model shows a fairly high level of Greek fluency, but it is less effective than larger-class and commercial models at reasoning and instruction following, and is more likely to experience hallucinations. 

In the future, our evaluation benchmarks should include more original Greek LLM datasets that are not the result of machine translation and post-editing. These datasets will help minimize the effect of machine translation on evaluation results and also better reflect the target language and culture.

\section{Risks and ethical considerations}\label{sec:ethical}

To mitigate potential risks, we took several steps to ensure the data used for training did not contain personally identifiable information, offensive, or otherwise inappropriate content. We sourced data from publicly available, licensed, or open-access datasets, ensuring compliance with their respective policies and any flagged data points were excluded. We did not collect data from private communications or data sources that could contain personally identifiable information. We have also given special care to align our model's responses with safety guidelines followed by manual reviews. 

We recognize that these measures are not a substitute for more thorough evaluation protocols, as we have not performed a systematic evaluation against LLM risks including risks related to discrimination, hate speech and exclusion, information hazards, and misinformation harms \citep{Wedidinger_etal_2022}. Moving forward, we aim to contribute towards addressing these limitations by promoting the development of Greek-language evaluation resources for LLM risks. This will enable more robust and contextually appropriate assessments of ethical risks in future models.

\section*{Acknowledgements}\label{sec:acknowledgements}

 The authors wish to thank AWS and GRNET, especially Nikiforos Botis and Panos Louridas, for their ongoing support and helping us attain the required training infrastructure. We express our sincere gratitude to all members of the Institute for Language and Speech Processing, Athena RC, for their unwavering support of this project. Special thanks to Maria Giagkou, Olga Yannoutsou and Penny Kyriazi for their help in translating various evaluation datasets and their efforts in the internal evaluations of the models.

\bibliography{krikri}

\newpage 
\onecolumn
\appendix

\section{Technical Details}\label{sec:appendix:technical_details}

\subsection{Pretraining Data Mix}\label{sec:appendix:data_mix}

The pretraining data mix (Table \ref{table:pretraining-data}) contains the following four components:

\begin{enumerate}
\item \textbf{Greek Texts (56.7B tokens):} The Greek part of the dataset was sourced from publicly available resources spanning a wide range of domains and sources: Wikipedia, ELRC-SHARE \cite{losch2021collection}, EUR-LEX \& MultiEUR-LEX \cite{chalkidis-etal-2019-large, chalkidis-etal-2021-multieurlex}, MaCoCu \cite{non-etal-2022-macocu}, CLARIN-EL \citep{GavriilidouEtAl2023}, EMEA\footnote{\url{https://www.ema.europa.eu/}}, parliamentary proceedings \cite{erjavec2023parlamint}, \footnote{\url{https://www.gutenberg.org/}}, governmental and legal documents from the Greek Government Gazette via the National Printing House\footnote{\url{https://et.gr/}}, the Permanent Greek Legislation Code -- Raptarchis dataset\footnote{\url{https://huggingface.co/datasets/AI-team-UoA/greek_legal_code}} \cite{papaloukas-etal-2021-glc}, Greek School Books\footnote{\url{https://ebooks.edu.gr/ebooks/}}, the Kallipos initiative of Greek open academic textbooks\footnote{\url{https://kallipos.gr/en/homepage/}}, full texts from publicly available articles, theses, and dissertations from academic repositories and the National Documentation Center\footnote{\url{https://www.ekt.gr/en}}, as well as pre-filtered resources originally compiled from the web, such as CulturaX \cite{nguyen2023culturax} and CulturaY\footnote{\url{https://huggingface.co/datasets/ontocord/CulturaY}}. In addition to Modern Greek, we incorporated a significant amount of Ancient Greek texts into our training corpus from Wikisource, school books, web pages, and Project Gutenberg\footnote{\url{https://www.gutenberg.org/}}, which provides freely available Ancient Greek texts, including classical literature and historical documents. By including Ancient Greek data, we ensured that Llama-Krikri-8B is able process polytonic Greek and engage with historical texts effectively. This enhances the model’s utility for classical studies, historical research, and philological applications.
\item \textbf{English Texts (21B tokens):} A subset of high-quality English data was mixed into the training corpus. This subset ensures that the model is continually trained on English data, and is drawn from sources that were also used for the Greek data, such as Wikipedia, Wikisource, Project Gutenberg (post-1900), EUR-LEX, EMEA, Greek academic repositories, etc. We also utilized additional English texts originating from abstracts and full texts of academic records found on multiple scientific repositories \cite{roussis2022scipar,roussis2024enhancing}, ECtHR-PCR \cite{tyss2024ecthr}, and pre-filtered datasets from TxT360 \cite{txt360data2024}, like ArXiv, S2ORC \cite{lo2020s2orc},  and PubMed Central \cite{roberts2001pubmed}. By incorporating diverse and high-quality English texts, we mitigate the risk of catastrophic forgetting.
\item \textbf{Parallel Data (5.5B tokens):} We compiled a diverse parallel corpus with language pairs covering multiple languages: Greek, English, French, Portuguese, German, Spanish, and Italian. The decision to add parallel data which covers other European languages (i.e., German, French, Italian, Portuguese, and Spanish) is informed from the languages that have been included in the multilingual instruction tuning of the original Llama-3.1 \cite{grattafiori2024llama3herdmodels}. We utilized resources such as SciPar \cite{roussis2022scipar}, MediSys \cite{roussis2022constructing}, MultiEUR-LEX \cite{chalkidis-etal-2021-multieurlex}, Europarl, TED Talk transcripts \cite{qi-EtAl:2018:N18-2}, and other sources with sentence pairs such as ELRC-SHARE \cite{losch2021collection} \& OPUS \cite{tiedemann2012parallel}. Our data include parallel documents and sentence pairs randomly sampled for each translation direction, e.g., EN-EL/EL-EN and EN-DE/DE-EN, as well as augmented training examples with concatenated parallel content across multiple languages (e.g., a Greek text followed by its English, German, and Spanish translations with appropriate prompt templates). 
The addition of these documents has a twofold effect. It has been shown that parallel data boosts translation performance \cite{alves2024tower,martins2024eurollm}, while limited empirical evidence indicates that pretrained LLMs process multilingual queries by first translating the content into English, utilizing their English knowledge to answer the query and then translate the answer back to the original language\cite{zhao2024large}.
\item \textbf{Code and Math (7.8B tokens):} We also integrated datasets containing text with code and mathematics, leveraging Stack Overflow\footnote{\url{https://huggingface.co/datasets/code-rag-bench/stackoverflow-posts}}, Python-Edu which is a subset of the SmolLM corpus~\cite{allal2025smollm2smolgoesbig} originating from The Stack V2 dataset \cite{lozhkov2024starcoder} and having been scored with an educational code classifier, and the AutoMathText dataset \cite{zhang2024automathtext}, which is a collection of math-related documents originating from web data, papers on arXiv, and code/notebooks on GitHub. AutoMathText has undergone an automatic selection process using Qwen-72B \cite{bai2023qwen} for relevancy to the mathematical domain and the educational value of each document. Code and Mathematics data, although not specific to Greek, were included to preserve and enhance the model's ability to handle coding tasks, math problems and formal language. Maintaining these capabilities broadens the utility of Llama-Krikri beyond pure language tasks.
\end{enumerate}

\subsection{Pretraining Data Cleaning Pipelines}\label{sec:appendix:data_cleaning}

Our filtering processes began with format standardization in order to facilitate uniform processing across multiple heterogeneous datasets. We converted all textual content from various formats (e.g., PDF, HTML, plain text, etc.) into JSONL containing both the document text and relevant metadata such as identified language, word count, and source information (including source URLs).

For PDF documents such as academic records and laws, we implemented a specialized pipeline which integrated Marker\footnote{\url{https://github.com/VikParuchuri/marker}} for extraction and conversion into Markdown files, as it exhibits strong performance for Greek texts. Subsequently, the pipeline included language identification using FastText \cite{joulin2017bag}, removal of markdown artifacts, and removal of lines with characters outside Unicode ranges for Greek, Latin, and other common and scientific symbols. Furthermore, we utilized document structure metrics (Marker also extracts various structural metadata) as quality indicators, such as the ratio of tables to pages and the fraction of removed lines in disallowed scripts.

Our main filtering pipeline used sequential rule-based and statistical filters to remove outlier documents across all data sources.  First, we implemented URL-based filtering by removing content from several blacklisted domains known to contain low-quality or problematic content. This was particularly effective for web-crawled datasets like CulturaX \cite{nguyen2023culturax} where relevant metadata was available. We then applied a set of minimal content-quality filters:
\begin{itemize}
\item Removal of documents containing multiple instances of profane or inappropriate terms from a curated list of Greek bad words
\item Removal of short documents based on character and word counts
\item Removal of documents containing multiple substrings like "lorem ipsum" which are indicative of content with low educational value
\item Removal of documents containing extremely long words (>60 characters)
\item Removal of documents with mean word length outside specified values.
\item Removal of documents with a high fraction of non-alphanumeric characters.
\end{itemize}

Parallel datasets were filtered using a different pipeline featuring various steps from previous work \cite{papavassiliou2018ilsp, roussis2022arc, roussis2024enhancing} which include: (a) rule-based filters, such as length ratio, language identification verification, and (b) model-based alignment quality scores using tools like LASER \cite{artetxe2018margin,artetxe2019massively}, BiCleaner AI \cite{zaragoza-bernabeu-etal-2022-bicleaner}, and CometKiwi \cite{rei-etal-2022-cometkiwi}.

Additionally, in order to mitigate privacy concerns and protect sensitive information, we systematically identified and anonymized personally identifiable information (PII) with the use of regular expressions. In particular, we aimed to detect and replace e-mail addresses with a generic placeholder (``email@example.gr'') and mask IP addresses (replacing them with 0.0.0.0).

Finally, for Greek, English, and Mathematics/Code datasets we implemented intra-dataset deduplication, as well as cross-dataset (global) deduplication. We utilized MinHashLSH near-deduplication \cite{broder1997resemblance, leskovec2020mining} with 5-gram subsets, a MinHash signature of 128, and a Jaccard similarity threshold of 0.8, following parameter choices similar to those used in other works \cite{nguyen2023culturax, voukoutis2024}. 

It should be noted that deduplication has consistently been shown to lead to higher performance, reduced training costs, as well as reduced model memorization; thus indirectly protecting sensitive information \cite{lee2022deduplicating,carlini2023quantifying,grattafiori2024llama3herdmodels,albalak2024survey}. However, global deduplication may also remove documents of high quality and actually hurt performance \cite{txt360data2024,penedo2024fineweb}. For this reason, we decided to upsample datasets of specific sources with important content, which leads to higher relative importance and also higher memorization \cite{carlini2023quantifying,txt360data2024}.

Regarding the deduplication of parallel datasets, we followed a different approach. All sentence pairs were normalized and cleaned, by converting them to lowercase and removing digits, punctuation. Pairs were then deduplicated based on the existance of either the source or target within the same dataset, thus ensuring that no sentence can be found multiple times in each parallel dataset \cite{roussis2022arc,roussis2024enhancing}.

Table~\ref{table:pretraining-data} summarizes the composition of the filtered and deduplicated pretraining corpus. In total, our collected dataset comprises roughly 91B tokens, of which 62.3\% is Greek text. For the final training curriculum, we upsampled parts of the corpus to effectively train on an equivalent of 110B tokens. Upsampling was used to give higher relative importance to certain underrepresented but valuable segments and it also leads to higher memorization of important content \cite{carlini2023quantifying,txt360data2024}. For example, we assigned a slightly higher weight to datasets with long-context documents, Wikipedia-like sources, dialogue data, multi-parallel documents, and to certain important domains, such as legal, scientific, and medical. The decision to include a significant amount of English and parallel data (23.1\% and 6\% of tokens, respectively) was guided by prior work \cite{voukoutis2024} showing that mixed-language training can help retain the base model's general knowledge and prevent catastrophic forgetting.

\subsection{Tokenizer Expansion}\label{sec:appendix:tokenizer-expansion-details}

The tokenizer and embeddings expansion process involved the following steps:

\begin{itemize}
\item \textbf{Data Acquisition:}
We acquired data by collecting sentences from high-quality sources of our pretraining mix in five domains: 
\begin{enumerate}
    \item \textbf{General domain} which reuses a sampled portion of the data used to train the tokenizer of Meltemi \cite{voukoutis2024} and covers diverse domains,
    \item \textbf{Legal domain} which includes legal texts.
    \item \textbf{Scientific domain} which uses publicly available articles, theses, and dissertations.
    \item \textbf{Literature domain} from public-domain literature, poetry, and other original writings across various variants of Greek (e.g., Koine Greek, Medieval Greek, Modern Greek, etc.),
    \item \textbf{Ancient Greek} which contains texts only in Ancient Greek.
\end{enumerate}

\item \textbf{Filtering and Preprocessing:}
Each dataset underwent sequential processing and filtering including language identification verification with FastText \cite{joulin2017bag}, application of regular expressions to remove URLs and other anomalies, symbol-to-word ratio filtering to remove outliers, and NFC normalization. We then performed sentence-level exact deduplication within each individual dataset. To ensure text quality, we applied fluency scoring using Monocleaner \cite{prompsit:2018:WMT} which leverages a 7-gram KenLM model for Greek, and setting a score threshold of 0.3 for non-polytonic text and 0.7 for other texts in Modern Greek.

\item \textbf{Creation of Train and Test Sets:}
For the tokenizer training and test sets creation, we sampled 50\% of the sentences from each source and divided it into train and test splits (80\%--20\%).
    
\item \textbf{Domain-specific Token Allocation:}
New tokens were added sequentially for each domain until tokenizer fertility for this domain remained relatively stable, with most of the tokens being allocated to the General domain. This approach ensured that common Modern Greek patters receive the largest coverage, while specialized terminology and older Greek variants are adequately represented.

\end{itemize}

In Table \ref{table:tokenizers_domain}, we list the token allocation per domain, as well as the tokenizer fertilities of Llama-3.1-8B and Llama-Krikri-8B for each of the five domains for which we created test sets during the tokenizer extention process. We can observe that tokenizer fertility has dropped significantly across all domains, with Ancient Greek benefiting from the highest reduction (-1.62) with just 492 added tokens, whereas fertility for Legal texts has almost halved (1.54 from 2.82) after the integration of a significant number of tokens (4,000).

\begin{table*}[!ht]
  \centering
  \small
\begin{tabular}{lcccc}
\hline
\textbf{Domain} & \textbf{Added Tokens} & \textbf{Llama-3.1-8B} & \textbf{Llama-Krikri-8B} & \textbf{Δ Fertility} \\
\hline
General & 15,000 & 2.65 & \textbf{1.59} & -1.06 \\
Legal & 4,000 & 2.82 & \textbf{1.54} & -1.28 \\
Scientific & 1,000 & 2.91 & \textbf{1.73} & -1.18 \\
Literature & 500 & 2.90 & \textbf{1.89} & -1.01 \\
Ancient Greek & 492 & 3.77 & \textbf{2.15} & -1.62 \\ 
\hline
\textbf{Total} & \textbf{20,992} & -- & -- & -- \\
\hline
\end{tabular}
\caption{Domain-specific token allocation and fertility comparison}
\label{table:tokenizers_domain}
\end{table*}

\subsection{Greek Embeddings Expansion \& Training}\label{sec:appendix:embedding_training}

Following the development of the Llama-Krikri-8B tokenizer with 149,248 tokens, containing the initial Llama 3 reserved tokens to reach a vocabulary size divisible by 256 for computational efficiency, we expanded the embeddings and lm head layer of our Llama 3 based model to match this vocabulary size. 

The newly initialized embeddings and lm head weights for each new token were initialized by averaging the embeddings (or lm head weights respectively) associated with the tokens retrieved by the initial Llama-3.1 tokenizer when fed with that token. This provided us with a smoother initial Greek embeddings and lm head training phase of the frozen model.

Prior to full-scale pretraining, we performed an initial adjustment of the newly introduced token embeddings with the configuration described in Table \ref{table:greek_embeddings_training}

\begin{table}[h]
  \centering
  \small
\begin{tabular}{ll}
\hline
\textbf{Parameter} & \textbf{Value} \\
\hline
Base model & Llama-3.1-8B-Base \\
Training data & 5B tokens \\
Training steps & 5,000 steps \\
Learning rate & $1e-3$ with cosine decay \\
Batch size & 2M tokens \\
Optimizer & AdamW \\
& $\beta_1 = 0.9$, $\beta_2 = 0.95$, $\epsilon = 10^{-5}$ \\
Weight decay & 0.1 \\
\hline
\end{tabular}
\caption{Greek embeddings training hyperparameters}
\label{table:greek_embeddings_training}
\end{table}

\subsection{Continual Pretraining Process}\label{sec:appendix:pretrain_process}

The continual pretraining phase was conducted over ~50 days on two AWS p5e.48xlarge instances (8× NVIDIA H200 GPUs). Throughout all training phases, we employed bf16 mixed precision to optimize memory usage and computational efficiency.

We employed a mixed-curriculum strategy during continual pretraining, alternating between chunks of predominantly Greek text and chunks of other supporting data in a round-robin fashion. Early in training, we emphasized simpler and higher-quality texts in Greek along with a higher proportion of English data. As training progressed, we introduced more diverse and complex Greek sources and increased the proportion of Greek relative to English per batch. This approach is akin to mini-batches containing a blend of languages, ensuring that every training epoch preserved a presence of English and technical content.

The hyperparameters for both the continual pretraining and annealing phases are presented in Table \ref{table:pretraining_hyperparams}.



\begin{table*}[ht]
  \centering
  \small
\begin{tabular}{lll}
\hline
\textbf{Parameter} & \textbf{Continual Pretraining} & \textbf{Annealing Phase} \\
\hline
Training corpus & 110B tokens (upsampled) & 3.5B + 189M synthetic QA tokens \\
Context length & 128K tokens & 128K tokens \\
Learning rate & Cosine annealing with linear warm-up & Linear decay from $3.0\times 10^{-5}$ to 0 \\
& over 1\% of steps from 0 to $3.0\times 10^{-4}$, & \\
& decay to $3.0\times 10^{-5}$ over 90\% of steps & \\
Batch size & $\sim$16M tokens & $\sim$16M tokens \\
Optimizer & AdamW ($\beta_1 = 0.9$, $\beta_2 = 0.95$, $\epsilon = 10^{-5}$) & AdamW ($\beta_1 = 0.9$, $\beta_2 = 0.95$, $\epsilon = 10^{-5}$) \\
Gradient clipping & 1.0 & 1.0 \\
Weight decay & lr $\times$ 0.1 & lr $\times$ 0.1 \\
Parallelization & DeepSpeed Zero 3 & DeepSpeed Zero 3 \\
\hline
\end{tabular}
\caption{Hyperparameters for continual pretraining and annealing phases}
\label{table:pretraining_hyperparams}
\end{table*}

\subsection{Annealing Phase}\label{sec:appendix:annealing}

For the annealing phase, we conducted experiments with two dataset configurations:

\begin{itemize}
    \item A curated dataset of 3.5B tokens from high-quality sources across all subcorpora
    \item The same dataset supplemented with 189M tokens of synthetic question-thinking-answer data
\end{itemize}

Table \ref{table:annealing_appendix} shows the results of these experiments on the Greek and English benchmarks that were used in subsection \ref{sec:evaluation_base} for base models. We can observe that continual pretraining improved Greek (+8.7) but reduced English (-4) vs Llama-3.1. Annealing with curated data gave modest gains. Most notably, adding synthetic QA significantly improved Greek (+2.1 vs continual pretraining) and enhanced English beyond original Llama-3.1 (+0.8).

In the annealing phase, we utilized Liger\cite{hsu2024ligerkernelefficienttriton} RMSNorm, RoPE, SwiGLU MLP and fused cross-entropy kernels to achieve lower memory consumption and higher training speeds, along with the hyperparameters presented in Table \ref{table:pretraining_hyperparams}.


\begin{table}[h]
  \centering
  \small
\begin{tabular}{lcc}
\hline
\textbf{Training Stage} & \textbf{Avg. Greek} & \textbf{Avg. English}\\
\hline
Llama-3.1-8B & 48.7 & 66.2 \\
+ Continual Pretraining & 57.4 & 62.2 \\
+ Curated Corpora & 58.0 & 63.4 \\
+ Synthetic QA Dataset & 59.5 & 67.0 \\
\hline
\end{tabular}
\caption{Average performance across training stages on Greek and English benchmarks}
\label{table:annealing_appendix}
\end{table}

\begin{table*}[ht]
  \centering
  \small
\begin{tabular}{lll}
\hline
\textbf{Parameter} & \textbf{SFT (Stages 1 \& 2)} & \textbf{DPO} \\
\hline
Training data & 856K pairs (Stage 1), 638K pairs (Stage 2) & 92K preference triplets \\
Learning rate & Linear decay from $5.0\times 10^{-6}$ to 0 with 3\% warmup & Linear decay from $2.5\times 10^{-7}$ to 0 over 2 epochs \\
Batch size & 128 & 32 \\
Optimizer & AdamW ($\beta_1 = 0.9$, $\beta_2 = 0.999$, $\epsilon = 10^{-5}$) & AdamW ($\beta_1 = 0.9$, $\beta_2 = 0.999$, $\epsilon = 10^{-5}$) \\
Gradient clipping & 1.0 & 1.0 \\
Weight decay & lr $\times$ 0.1 & lr $\times$ 0.1 \\
Special features & Loss masking on prompt tokens & Length normalization \\
\hline
\end{tabular}
\caption{Hyperparameters for supervised fine-tuning and direct preference optimization}
\label{table:instruction_tuning_hyperparams}
\end{table*}

\subsection{Instruction Tuning and Alignment}\label{sec:appendix:instruction_tuning}

The SFT phase was implemented in two distinct stages with progressively increasing data quality and complexity, so as to incrementally improve the model's instruction-following capabilities. In Stage 1, we utilized extended datasets including translated ones, while Stage 2 emphasized higher-quality data with more complex reasoning patterns. For both stages, training used AdamW~\cite{adamw2017} and we masked the cross-entropy loss of prompt tokens so the model would only learn based on the completions.

For the DPO phase, we applied length normalization as described by \citet{rafailov2024directpreferenceoptimizationlanguage}. At its core the DPO objective maximizes the likelihood of preferred responses while minimizing the likelihood of dispreferred ones, under a regularization term, while the length normalization term ensures the model is not biased toward longer responses. This approach helped balance verbosity and brevity, optimizing for overall usefulness. The final model checkpoint was selected based on the largest margin between preferred and dispreferred completions during validation.
Training used the AdamW~\cite{adamw2017} optimizer, while Liger\citep{hsu2024ligerkernelefficienttriton} RMSNorm, RoPE, SwiGLU MLP, fused cross-entropy kernels, and fused DPO loss kernels were utilized to achieve lower memory consumption and higher training speeds during SFT and DPO. The hyperparameters for both are listed in Table \ref{table:instruction_tuning_hyperparams}.

\newpage 
\onecolumn

\section{Examples of queries and responses}

\begin{mytbox}[]{Programming: Python Fibonacci function}
\textbf{User:} Γράψε μια συνάρτηση Python που υπολογίζει την ακολουθία fibonacci. 

\textbf{Krikri:}  Ακολουθεί η υλοποίηση της αναδρομικής συνάρτησης Python για τον υπολογισμό των αριθμών Fibonacci:

\begin{minted}[breaklines]{python}
def fibonacci(n):
    """
    Υπολογίζει τον n-οστό αριθμό Fibonacci.
    
    Args:
        n (int): θετικός ακέραιος. n >= 0 (0-indexed προσέγγιση)
    
    Returns:
        int: ο n-οστός αριθμός Fibonacci
    
    Raises:
        ValueError: αν n είναι αρνητικός αριθμός
    """
    # Έλεγχος εγκυρότητας εισόδου
    if not isinstance(n, int) or n < 0:
        raise ValueError("Η είσοδος πρέπει να είναι μη αρνητικός ακέραιος")
    
    # Βάση αναδρομής: 
    # F(0) = 0, F(1) = 1 (κλασικοί ορισμοί)
    if n in (0, 1):
        return n
    
    # Αναδρομικός τύπος: F(n) = F(n-1) + F(n-2)
    return fibonacci(n-1) + fibonacci(n-2)
\end{minted}

\textbf{Παράδειγμα χρήσης:}

\begin{minted}{python}
# Υπολογισμός μερικών όρων της 
# ακολουθίας Fibonacci
for i in range(10):
    print(f"F({i}) = {fibonacci(i)}")
    
# Εκτύπωση αποτελεσμάτων
# F(0) = 0
# F(1) = 1
# F(2) = 1
# F(3) = 2
# F(4) = 3
# F(5) = 5
# F(6) = 8
# F(7) = 13
# F(8) = 21
# F(9) = 34
\end{minted}
\end{mytbox}

\begin{mytbox}[]{Translation: Ancient to Modern Greek}
\textbf{User:} {\selectlanguage{greek}Μετάφρασέ μου στα Νέα ελληνικά το κείμενο: \\
``Οἱ μὲν δὴ Μήλιοι τοσαῦτα ἀπεκρίναντο· οἱ δὲ Ἀθηναῖοι διαλυόμενοι ἤδη ἐκ τῶν λόγων ἔφασαν «ἀλλ’ οὖν μόνοι γε ἀπὸ τούτων τῶν βουλευμάτων, ὡς ἡμῖν δοκεῖτε, τὰ μὲν μέλλοντα τῶν ὁρωμένων σαφέστερα κρίνετε, τὰ δὲ ἀφανῆ τῷ βούλεσθαι ὡς γιγνόμενα ἤδη θεᾶσθε, καὶ Λακεδαιμονίοις καὶ τύχῃ καὶ ἐλπίσι πλεῖστον δὴ παραβεβλημένοι καὶ πιστεύσαντες πλεῖστον καὶ σφαλήσεσθε.»''}

\textbf{Krikri:} {\selectlanguage{greek}Ακολουθεί η μετάφραση του κειμένου στα Νέα Ελληνικά:

Οι Μήλιοι απάντησαν με αυτά τα λόγια. Οι Αθηναίοι, καθώς αποχωρούσαν από τη συζήτηση, είπαν: ``Όμως εσείς είστε οι μόνοι που από αυτές τις συζητήσεις συμπεραίνετε σωστά τόσο τα ορατά όσο και τα αόρατα, σαν να έχουν ήδη συμβεί. Είστε υπερβολικά αισιόδοξοι και έχετε εναποθέσει τις ελπίδες σας στους Λακεδαιμονίους, στην τύχη και στις προσδοκίες σας. Και όσο περισσότερο έχετε πιστέψει σε αυτά, τόσο περισσότερο θα απογοητευτείτε.''}
\end{mytbox}

\begin{mytbox}[]{Translation to JSON Output: Greek to Multiple Languages}
\textbf{User:} {\selectlanguage{greek}Μετάφρασέ μου το παρακάτω κείμενο από τα νέα ελληνικά στα αγγλικά, στα ισπανικά και στα γαλλικά. Δώσε μου τις μεταφράσεις σε \selectlanguage{english}JSON. \\
\\
Σύμφωνα με την οδηγία 2009/28/ΕΚ του Ευρωπαϊκού Κοινοβουλίου, ως ενέργεια από ανανεώσιμες μη ορυκτές πηγές θεωρείται η αιολική, ηλιακή, αεροθερμική, γεωθερμική, υδροθερμική και ενέργεια των ωκεανών, υδροηλεκτρική, από βιομάζα, από τα εκλυόμενα στους χώρους υγειονομικής ταφής αέρια, από αέρια μονάδων επεξεργασίας λυμάτων και από βιοαέρια.}

\textbf{Krikri:} \texttt{\\
\{ \\
"en": "According to Directive 2009/28/EC of the European Parliament, energy from renewable non-fossil sources includes wind, solar, aerothermal, geothermal, hydrothermal and ocean energy, hydropower, biomass, landfill gas, sewage treatment plant gas and biogases.", \\
"es": "De conformidad con la Directiva 2009/28/CE del Parlamento Europeo, la energía procedente de fuentes renovables no fósiles incluye la eólica, solar, aerotérmica, geotérmica, hidrotérmica y la energía del océano, hidroeléctrica, procedente de la biomasa, de los gases de vertedero, de los gases de estación depuradora de aguas residuales y de los biogás.", \\
"fr": "Conformément à la directive 2009/28/CE du Parlement européen, les énergies renouvelables non fossiles comprennent l'éolien, le solaire, l'aérothermique, la géothermie, l'hydrothermie et l'énergie océanique, l'hydroélectricité, la biomasse, les gaz de décharge, les gaz de station d'épuration et les biogaz." \\
\}}
\end{mytbox} 

\begin{mytbox}[]{Summarization: Basketball game}
\textbf{User:} Μπορείς να δημιουργήσεις μια περίληψη 5 προτάσεων για τον παρακάτω αγώνα μπάσκετ: (User provides the text of an article describing a 2025 basketball game)

\textbf{Krikri:} {\selectlanguage{greek}Ο Ολυμπιακός νίκησε την Μπαρτσελόνα με 90-88 σε ένα συναρπαστικό παιχνίδι, χάρη σε ένα νικητήριο καλάθι του Φουρνιέ στην εκπνοή. Ο Βεζένκοφ και ο Γκος σημείωσαν από 24 και 20 πόντους αντίστοιχα για τους νικητές, ενώ ο Μετού ήταν ο κορυφαίος των Καταλανών με 16 πόντους. Η ομάδα του Μπαρτζώκα παρέμεινε στην κορυφή της Ευρωλίγκας μετά από αυτή τη σημαντική νίκη.}
\end{mytbox}

\begin{mytbox}[]{History: Population exchange between Greece and Turkey}
\textbf{User:} {\selectlanguage{greek}Συνόψισε σε μία παράγραφο την Ελληνοτουρκική ανταλλαγή πληθυσμών του 1923}

\textbf{Krikri:} {\selectlanguage{greek}Η Ελληνοτουρκική ανταλλαγή πληθυσμών του 1923, που προβλέφθηκε από τη Συνθήκη της Λωζάνης, αποτέλεσε μια μαζική μετακίνηση πληθυσμών μεταξύ Ελλάδας και Τουρκίας με στόχο την ομογενοποίηση των εθνών-κρατών. Περίπου 1,5 εκατομμύριο Έλληνες χριστιανοί, κυρίως από την Ανατολική Θράκη, την Ιωνία, τον Πόντο και την Καππαδοκία, μετακινήθηκαν στην Ελλάδα, ενώ αντίστοιχα 500.000 Μουσουλμάνοι (κυρίως Τούρκοι αλλά και Πομάκοι και Αθίγγανοι) μετακινήθηκαν από την Ελλάδα στην Τουρκία. Αυτή η ανταλλαγή, παρά τις τραγικές της συνέπειες για τους ανθρώπους που εκτοπίστηκαν, αποτέλεσε ένα από τα σημαντικότερα γεγονότα που διαμόρφωσαν τη σύγχρονη δημογραφική και εθνοτική σύνθεση των δύο χωρών, οδηγώντας σε μια πιο ομοιογενή εθνική ταυτότητα για κάθε κράτος. Η διαδικασία αυτή ολοκληρώθηκε το 1924 και θεωρείται ορόσημο στις ελληνοτουρκικές σχέσεις.}
\end{mytbox}

\begin{mytbox}[]{High school biology: The process of photosynthesis}
\textbf{User:} Δώσε μια σύνοψη της διαδικασίας της φωτοσύνθεσης σε μία παράγραφο.

\textbf{Krikri:} {\selectlanguage{greek}Η φωτοσύνθεση είναι η διαδικασία με την οποία τα φυτά, χρησιμοποιώντας την ηλιακή ενέργεια, μετατρέπουν το διοξείδιο του άνθρακα (CO2) και το νερό (H2O) σε γλυκόζη (C6H12O6) και οξυγόνο (O2). Συγκεκριμένα, η χλωροφύλλη στα φύλλα απορροφά φως, διασπά το νερό απελευθερώνοντας οξυγόνο, και χρησιμοποιώντας το υδρογόνο συνθέτει γλυκόζη από το διοξείδιο του άνθρακα. Η παραγόμενη γλυκόζη αποτελεί πηγή ενέργειας για το φυτό, ενώ το οξυγόνο απελευθερώνεται στην ατμόσφαιρα ως σημαντικό παραπροϊόν, υποστηρίζοντας την επιβίωση των αερόβιων οργανισμών.}
\end{mytbox}

\begin{mytbox}[]{Chain-of-Thought Reasoning: Math problem}
\textbf{System:} (Appropriate system prompt used to enable thinking mode)

\textbf{User:} Η Αλίκη έχει ένα καλάθι με μήλα. Δίνει τα 2/5 των μήλων στην αδερφή της και το 1/5 στον αδερφό της. Τελικά της μένουν 6 μήλα. Πόσα μήλα είχε αρχικά στο καλάθι;

\textbf{Krikri:} \begin{Verbatim}[commandchars=\\\{\}]
<thinking>
Ας λύσουμε αυτό το πρόβλημα βήμα προς βήμα:

1. Έστω x ο αρχικός αριθμός μήλων που είχε η Αλίκη
2. Γνωρίζουμε ότι:
   * Δίνει 2/5 του x στην αδερφή της
   * Δίνει 1/5 του x στον αδερφό της
   * Μετά από όλα αυτά της μένουν 6 μήλα

3. Ας κάνουμε την εξίσωση:
   * Αρχικά είχε: x μήλα
   * Της μένουν: 6 μήλα
   * Άρα: x - (2/5)x - (1/5)x = 6
   * Απλοποιώντας: x - (3/5)x = 6
   * Συνδυάζοντας όρους: (2/5)x = 6
   * Πολλαπλασιάζοντας και τα δύο μέλη με 5/2:
   * x = 15

Ας επαληθεύσουμε:
* Αρχικά είχε 15 μήλα
* Δίνει: 2/5 * 15 = 6 μήλα στην αδερφή της
* Δίνει: 1/5 * 15 = 3 μήλα στον αδερφό της
* Μένουν: 15 - 6 - 3 = 6 μήλα
Η λύση είναι σωστή!
</thinking>

<output>
Η Αλίκη είχε αρχικά 15 μήλα στο καλάθι της.

Μπορούμε να το επαληθεύσουμε:
* Έδωσε 6 μήλα στην αδερφή της (2/5 του συνόλου)
* Έδωσε 3 μήλα στον αδερφό της (1/5 του συνόλου)
* Έτσι της έμειναν 6 μήλα (15 - 6 - 3 = 6)
</output>
\end{Verbatim}
\end{mytbox}

\end{document}